\documentclass[conference]{IEEEtran}
\IEEEoverridecommandlockouts
\usepackage{cite}
\usepackage{amsmath,amssymb,amsfonts}
\usepackage{algorithmic}
\usepackage{graphicx}
\usepackage{textcomp}
\usepackage{xcolor}
\usepackage{colortbl}
\usepackage{diagbox}
\usepackage[ruled,vlined]{algorithm2e}
 \usepackage{hyperref}
 \usepackage{bm}

\def\BibTeX{{\rm B\kern-.05em{\sc i\kern-.025em b}\kern-.08em
    T\kern-.1667em\lower.7ex\hbox{E}\kern-.125emX}}
\begin{document}

\title{High-dimensional Multivariate Time Series Forecasting in IoT Applications using Embedding Non-stationary Fuzzy Time Series\\
\thanks{This work has been supported by the Brazilian agencies (i) National Council for Scientific and Technological Development (CNPq); (ii) Coordination for the Improvement of Higher Education (CAPES) and (iii) Foundation for Research of the State of Minas Gerais (FAPEMIG, in Portuguese).}
\thanks{MINDS Laboratory -- \url{https://minds.eng.ufmg.br/}}
}
\author{\IEEEauthorblockN{Hugo Vinicius Bitencourt}
\IEEEauthorblockA{\textit{Machine Intelligence and Data Science Lab (MINDS)} \\
\textit{Graduate Program in Electrical Engineering}\\
\textit{Universidade Federal de Minas Gerais}\\
Av. Antônio Carlos 6627, 31270-901 \\ 
Belo Horizonte, MG, Brazil \\
hugovynicius@ufmg.br}
\and
\IEEEauthorblockN{Frederico Gadelha Guimarães}
\IEEEauthorblockA{\textit{Machine Intelligence and Data Science Lab (MINDS)} \\
\textit{Universidade Federal de Minas Gerais}\\
Av. Antônio Carlos 6627, 31270-901 \\
Belo Horizonte, MG, Brazil \\
fredericoguimaraes@ufmg.br}

}

\maketitle

\begin{abstract}
In Internet of things (IoT), data is continuously recorded from different data sources and devices can suffer faults in their embedded electronics, thus leading to a high-dimensional data sets and concept drift events. Therefore, methods that are capable of high-dimensional non-stationary time series are of great value in IoT applications. Fuzzy Time Series (FTS) models stand out as data-driven non-parametric models of easy implementation and high accuracy. Unfortunately, FTS encounters difficulties when dealing with data sets of many variables and scenarios with concept drift. We present a new approach to handle high-dimensional non-stationary time series, by projecting the original high-dimensional data into a low dimensional embedding space and using FTS approach. Combining these techniques enables a better representation of the complex content of non-stationary multivariate time series and accurate forecasts. Our model is able to explain 98\% of the variance and reach 11.52\% of RMSE, 2.68\% of MAE and 2.91\% of MAPE.

\end{abstract}


\section{Introduction}

Internet of Things (IoT) can impact on several aspects of everyday-life and behavior of potential users. Sensor nodes and actuators distributed in houses and offices can make our life more comfortable, for example: rooms heating can be adapted to our preferences and to the weather; domestic incidents can be avoided with appropriate monitoring and alarm systems; and energy can be saved by automatically switching off the electrical equipments when not needed. Sensor nodes can be used for factory automation, inventory management, and detection of liquid/gas leakages \cite{IoTSurvey} \cite{IoTSurvey2012} \cite{IoTSurvey2013}. 

The growth of IoT applications in Industry 4.0 and smart homes and the increasing availability of data storage has led to an enormous rising amount of data being produced in a streaming fashion. This data is arranged in the form of a time series. Unfortunately, sensors nodes may suffer from inevitable aging effects, or faults in their embedded electronics. Besides, the physical phenomena under monitoring can also evolve with time due to seasonality or meteorological changes \cite{Learning_Nonstationary_Environments}. 

These time series are characterized by intrinsic changes that modify the properties of the data generating process (i.e. non-stationary time series), then changing its underlying probability distribution over time. A non-stationary time series is defined in terms of its mean or variance (or both) varying over time and the changes can take several forms, a phenomenon known as “concept drift”. The "concept drift” may deteriorate the accuracy of model prediction over time, which requires permanent adaptation strategies.

In the context of IoT, data is continuously recorded from different data sources and each sensor produces a streaming time series, where each time series dimension represents the measurements recorded by a sensor node, thus leading to a high-dimensional time series. Formally, an IoT application with $M$ sensors generates an $M$-dimensional time series. Besides, high-dimensional streaming time series is one of the most common type of dataset in the big data.

Time series methods that are capable of handling high-dimensional non-stationary time series are of great value in IoT applications. The analysis of such datasets poses significant challenges, both from a statistical as well as from a numerical point of view.

Fuzzy Time Series (FTS) methods became attractive due to their easy implementation, low computational cost, forecast accuracy and model interpretability. However, as the dimensionality of time series increases, FTS methods notably lose their accuracy and simplicity. Since each variable has its own fuzzy sets and the number of rules in a multivariate FTS model is given by a combination of the fuzzy sets, the number of rules may grow exponentially with the number of variables. Therefore, there is noticeable gap in adopting FTS models for high-dimensional time series and scenarios with concept drift \cite{hyper_otim_fts}\cite{nsfts}.

To overcome this challenge, we present a new approach to predict high-dimensional non-stationary streaming data generated by sensors in IoT applications. We apply data embedding transformation and use FTS models. The embedding allows us to extract a new feature space that better represents the complex content of multivariate time series data for the subsequent forecasting task. This work stands out as one of the few methods presented in the literature of FTS models to address the problem of high-dimensional non-stationary time series. 

The rest of the paper is organized as follows . The related work, from both application and methodological point of views, are presented in Section~\ref{background}. In Section~\ref{emb-nsfts}, we describe in detail the proposed approach. Section~\ref{experiments} describes a case of study of a smart home application used to test our method. The results of the case study are presented and discussed in Section~\ref{results}. Section~\ref{conclusions} concludes the paper.    

\section{Background}
\label{background}

\subsection{Smart Homes}
\label{Smart homes}

Smart homes is one of the most popular IoT applications and sensor nodes have been used to collect data to analyse the behavior and proper uses of energy. Energy consumption prediction is very important for smart homes, since it helps reduce power consumption and provides better energy and cost savings. Several machine learning algorithms have been used for forecasting energy consumption using data collected from sensor nodes. 

Four prediction models (Multiple Linear Regression (MLR), Support Vector Machine with Radial Kernel (SVM-radial), Random Forest (RF) and Gradient Boosting Machines (GBM)) were implemented and evaluated in \cite{dataset_appliance} for the energy use of appliances in a low-energy house in Belgium. A Multilayer Perceptron (MLP) with four hidden layers and 512 neurons in each layer was also used to predict the same household appliance energy consumption in \cite{energy_prediction_mlp}. Extreme Random Forest (ERF), K-nearest neighbor (KNN) and LSTM were used to build forecasting models for the same appliances energy consumption problem in \cite{energy_prediction_lstm}. 

\subsection{Dimensionality Reduction}
\label{dimensionality_reduction}

There are several approaches for dealing with high dimensional data in the literature. Some of the major dimension reduction (embedding) techniques are feature selection and feature extraction. 
In feature selection, a subset of the original features is selected. On the other hand, in feature extraction, a set of new features are found through some mapping from the existing input variables. The mapping may be either linear or non-linear.

The goal of embedding by feature extraction is to learn a function $\gamma:\mathbb{R}^{M} \rightarrow \mathbb{R}^{K}$ which maps $M$-dimensional features measured (i.e. time series) over $T$ time steps into the reduced $K$-dimensional feature space with $K~\ll~M$. 

Principal Component Analysis (PCA) \cite{pca} is one of the most popular feature extraction approaches. PCA estimates the cross-correlation among the variables and extracts a reduced set of features which are linearly uncorrelated. The main limitation of PCA method is its ability to capture only linear correlation among variables.

In real word, it is common to find nonlinear correlation, then we can use a nonlinear PCA analysis named KPCA (Kernel Principal Component Analysis) \cite{kpca}. In KPCA, the idea of kernel function is used in order to handle nonlinear feature extraction by finding a suitable nonlinear mapping function $\Phi$, which is called kernel function. Gaussian radial basis function (RBF), Polynomial kernel and Sigmoid kernel are examples of kernel function that can be used in KPCA. 

\subsection{Fuzzy Time Series}
\label{fts}

The fundamentals of Fuzzy Time Series (FTS) were first proposed by Song and Chissom \cite{fts_paper} to handle unclear and imprecise knowledge in time series data. FTS is a representation of the time series using fuzzy sets as fundamental components, then conventional time series values are transformed to linguistic time series. Since the introduction of the FTS, several categories of FTS methods have been proposed, varying by their order ($ \Omega $) and time-variance. Order is the number of time-delays (lags) that are used in modeling the time series. The time variance defines whether the FTS model changes over time.

In the training procedure of an FTS model, the Universe of Discourse ($U$) is partitioned into intervals that are limited by the known bounds of $Y$, where $U = [\min(Y), \max(Y)]$. For each interval, a fuzzy set $A_i \in \tilde{A}$ is defined with its own membership function (MF) $\mu_{A_i}:\mathbb{R} \rightarrow [0,1]$, then a linguistic value is assigned to each fuzzy set and represents a region of $U$. The crisp time series $Y$ is mapped onto the fuzzified time series $F$, given the membership values to the fuzzy sets. Temporal patterns are created from $F$ according to the number of lags $ \Omega $. Each pattern represents a fuzzy rule called Fuzzy Logical Relationship (FLR) and they are grouped by their same precedents forming a Fuzzy Logical Relationship Group (FLRG). 

Once the FTS model is trained, it can be used to forecast new values. The crisp samples $y(t-\Omega), \ldots, y(t-1)$ are mapped onto the fuzzified values $f(t-\Omega), \ldots, f(t-1)$, where $f(t) = \mu_{A_i}(y(t)), \forall A_i \in \tilde{A} $, for $t = 1, \ldots, T$. The rules that match with the corresponding input are found. The FLRG whose precedent is equal to the input value is selected and the candidate fuzzy sets in its consequent are applied to estimate the forecast value. 

Non-stationary fuzzy sets (NSFS) were proposed by Garibaldi, Jaroszewski and Musikasuwan and Garibaldi and Ozen, respectively in \cite{nsfs_paper_1} and \cite{nsfs_paper_2}. They proposed a dynamic item which changes the membership function over time in the fuzzy sets. A NSFS is defined by the non-stationary membership function (NSMF) that considers time variations of the corresponding membership function, and the perturbation function which is the dynamic component responsible for changing the parameters of the MF, given some parameter set. 

We can use NSFS for non-stationary series forecasting problems, however NSFS is only suitable when the variance of the data distribution changes in a predictable way, limiting its performance to more complex changes, such as concept drift events. The Non-Stationary Fuzzy Time Series (NSFTS) \cite{nsfts} is an FTS method that is able to dynamically adapt its fuzzy sets to reflect the changes in the underlying stochastic processes based on the residual errors. The NSFTS model can handle non-stationary time series as well as scenarios with concept drift. Unfortunately, NSFTS is a uni-variate method, hence it cannot handle high dimensional multivariate time series.

\section{Embedding Non-stationary Fuzzy Time Series}
\label{emb-nsfts}

We extend the NSFTS (ENSFTS) in order to enable it to high dimensional multivariate time series, by applying embedding transformation with PCA and KPCA, then reducing the time series dimensionality and allowing efficient pattern discovery and induction of fuzzy rules.

\begin{figure*}[!h]
\begin{center}
 \includegraphics[width=16cm]{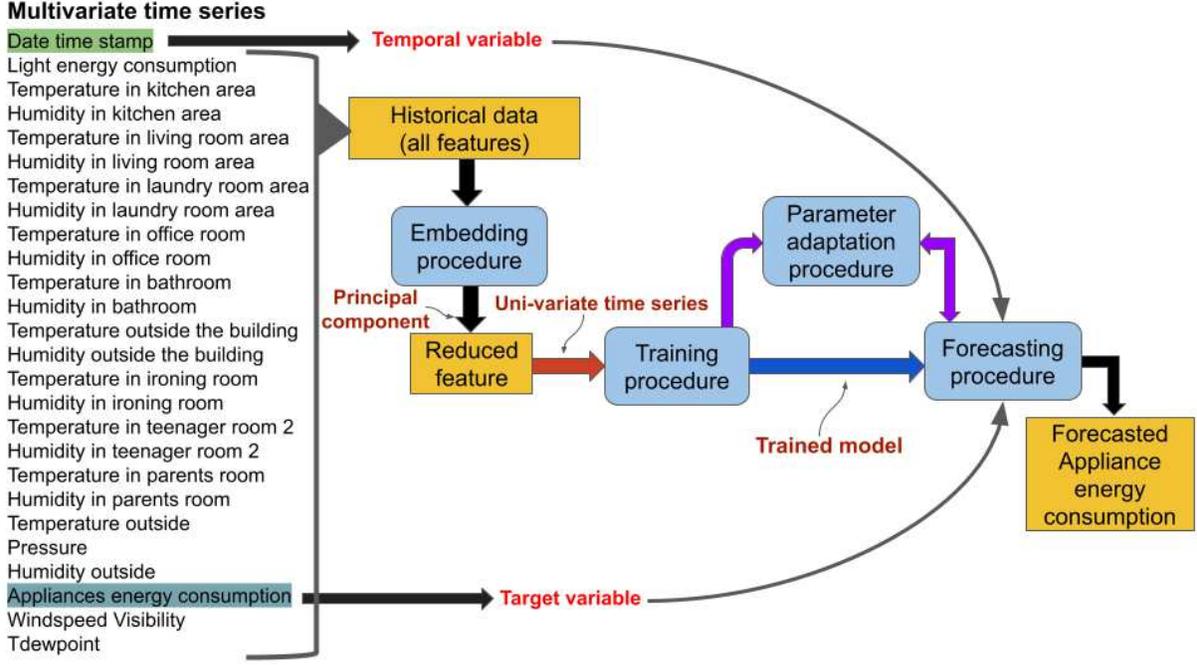}
  \caption{ENSFTS proposed method}
  \label{model}
\end{center}
\end{figure*}

The ENSFTS method is a data-driven and explainable method which is flexible and adaptable for many IoT applications. The proposed approach, depicted in Figure~\ref{model}, consists of embedding, training, parameter adpatations and forecasting procedures. 

In this work, the proposed approach aims to address the appliances energy consumption forecasting problem. The embedding algorithm is used to extract the main components that better represent the content of appliance energy consumption multivariate time series for the subsequent forecasting task. The procedures are detailed below.

\subsection{Embedding}
\label{embedding}

Regarding PCA implementation, we take the following steps. Given the multivariate time series $Y \in \mathbb{R}^{N\times M}$, we calculate the covariance matrix $C \in \mathbb{R}^{N \times M}$ extracting the first $K$ eigenvectors related to the largest eigenvalues, obtaining the matrix $Z \in \mathbb{R}^{N \times M}$ that is used to calculate the embedding feature $\gamma(x):Z^T  \cdot y $ where $y \in \mathbb{R}^{M}$. 

We take the following steps to implement the RBF kernel PCA. First, construct the kernel similarity matrix, according to equation 
\begin{equation} \label{eq:kernel}
    k({\textbf{x}_i},{\textbf{x}_j})=\exp({{- \gamma ||{\textbf{x}_i}-{\textbf{x}_j}||^2_{2}}})
\end{equation}
where $\gamma$ is the kernel coefficient. Second, since it is not guaranteed that the kernel matrix is centered, we compute the centered kernel matrix $\widetilde{K} = K - 1_N K - K 1_N + 1_N K 1_N$, where $1_N$ is an $N \times N$ matrix with all values equal to $1/N$. Third, solve the eigenvalue problem $\widetilde{K} \bm{\alpha}_i = \lambda_i N \bm{\alpha}_i$. Finally, obtain the eigenvectors of the centered kernel matrix that correspond to the largest eigenvalues. Those eigenvectors are the data points projected onto the respective principal components.

Therefore, we transform $M$ features of the data set to one feature and apply the NSFTS model, then reducing the multivariate time series to uni-variate time series. NSFTS combined with PCA and KPCA are named PCA-NSFTS and KPCA-NSFTS, respectively.  

\subsection{Training}
\label{training}

The training procedure generates a model that captures all the information in the embedding data, leaving a residual $E \sim N(0,1)$. Let the embedding time series $Y_{emb} \in \mathbb{R}^{1}$ and its individual instances $y_{emb}(t)$ $\in$ $Y_{emb}$ for $t = 0, 1,..., T$, the number of fuzzy sets $\kappa$ and the length of the residuals window $w_e$. 

Define $U=[lb, ub]$, where $lb = \min(Y_{emb}) - D_1$ and $ub = \max(Y_{emb})+ D_2$, with $D_1 = r \times |\min(Y_{emb})|$ and $D_2 = r \times | \max(Y_{emb})|$, $0 < r < 1$. The partitioning is oriented by the midpoints $c_i$ of each fuzzy set $A_i$, according to equation  
\begin{equation} \label{eq:midpoints}
    c_i = lb + i \times \frac{ub-lb}{\kappa-1} 
\end{equation}

For each interval, a fuzzy set is defined with its triangular membership function $\mu_{A_i}(y_{emb})$ 
$$
\mu_{A_i}(y_{emb})=\begin{cases}
			0, & \text{if $y$ $<$ $l$ or $y$ $>$ $u$}\\
            \frac{y-l_i}{c_i-l_i}, &  \text{if $l_i$ $\leq$ $y$ $\leq$ $c_i$} \\
            \frac{u_i-y}{u_i-c_i}, &  \text{if $c_i$ $\leq$ $y$ $\leq$ $u_i$} 
		 \end{cases}
$$

All fuzzy sets have a perturbation function $ \pi_i $ 
\begin{equation} \label{eq:pertubation}
    \pi(l,c,u,\delta,\rho) = \left\{ \frac{\rho}{2}-(l+\delta),c+\delta,\frac{\rho}{2}+(u+\delta) \right\} 
\end{equation}
where $\delta$ is the displacement of $A_i$ along $U$ and $\rho$ the scale responsible for changing the $A_i$ coverage area, either by stretching or contracting its shape. These parameters are initialized as zero.

The time series $Y_{emb}$ is then transformed into an FTS $F_{emb}$ and temporal patterns with the format $A_p \rightarrow A_c$ are extracted where $A_p$ is the precedent and $A_c$ the consequent, and both are related to $A_i$ with maximum membership. As explained before, each pattern represents a fuzzy rule and they are grouped by their same precedents. 

Finally, we compute the residuals by applying the forecasting procedure to the training set. The last $w_e$ items are forecasted in order to calculate the residuals as follow   

\begin{equation} \label{eq:pertubation}
    E = \left\{ e(t-w_e), e(t - (w_e-1),...,e(t)) \right\} 
\end{equation}
where $e(t) = y_{emb}(t) - \hat{y}_{emb}(t)$ and $\hat{y}_{emb}(t)$ is the predicted value.

\subsection{Parameter adaptation}

In the parameter adaptation, the mean and variance of the residuals are monitored and used to change the membership function. Let the residuals $E$, the forecast value $\hat{y}(t + 1)$ and its corresponding actual value $y(t + 1)$. The displacement parameter is updated from changes in its midpoints that are detected when $y(t)$ is outside the range of the U, according with the following conditions 
\begin{equation} \label{eq:displacement_update}
    \textit{IF}~(y(t) < lb)~~ \textit{THEN}~(d_l = lb - y(t))~~ \textit{ELSE}~(d_l = 0) 
\end{equation}
\begin{equation} \label{eq:displacement_update_2}
     \textit{IF}~(y(t) > ub)~~ \textit{THEN}~(d_u = y(t) - ub)~~ \textit{ELSE}~(d_u = 0)
\end{equation}

We compute the mean $\bar{E}$ and variance $\sigma_E$ of the residuals. These values are used to update position and length of the fuzzy sets. For each fuzzy set, the displacement $\delta_i$ is calculated according with the following equation 

\begin{equation} \label{eq:displacement}
    \delta_i = \bar{E} + \left(i\frac{r}{k+1}-d_{mp} \right) + \left(i\frac{2\sigma_E}{k-1}-\sigma_E \right) 
\end{equation}
where $r = d_u - d_l$ (displacement range) and $d_{mp} = r/2$ (displacement midpoint). 

Finally, we compute the scaling factors $\rho_i$ as follow: $\rho_i = |\delta_{i-1}-\delta_{i+1}|$. The new parameters values $\delta_i$ and  $\rho_i$ are used by the perturbation function. 

\subsection{Forecasting}

The forecasting procedure finds the rules that match a given fuzzified input and use them to calculate a numerical forecasting using non-stationary fuzzy sets perturbed by $\pi$. 

Given the target univariate time series $Y$ and its instances $y(t)$ for $t = 0, 1,..., T$, the following steps are taken to forecast $\hat{y}(t + 1)$. We calculate the membership grade $ \mu_{A_i} $ for each fuzzy set $A_i$ using the $MF$ with the parameters adapted by $\pi$, and we select the fuzzy sets $A_j$ where $\mu_{A_i}  \geq 0$. 

$A_j$ are the inputs for the rule base in order to match rules according to their precedent. The rule set is defined as $ S = \{A_j \rightarrow C_j ~|~\mu_{A_j}(y(t)) > 0\}$, where $C_j$ is the consequent of the rule.

The predicted value $\hat{y}(t + 1)$ is obtained as the weighted sum of the rule midpoints by their membership grades $\mu_{A_j}$, according to equation:

\begin{equation} \label{eq:sum_forecasting_fts}
    \hat{y}(t + 1) = \sum\limits_{A_j \rightarrow C_j \in S} \mu_{A_j}(y(t))\cdot mp(C_j)
\end{equation}

with $mp(C)$ determined as follows:
\begin{equation} \label{eq:midpoints_forecasting}
mp(C) = \frac{\sum\nolimits_{A_i \in C} c_{A_i}}{|C|}
\end{equation}

\section{Experiments}
\label{experiments}

\subsection{Case of study}
\label{case_study}

An important application of IoT in smart homes is the monitoring of appliances energy consumption. This importance is due to the fact that the correct monitoring of energy appliances can reduce power consumption and provides better energy and cost savings. 

As an example of the approach presented here, we use the data set of energy appliances presented in \cite{dataset_appliance}. The data set includes measurements of temperature and humidity collected by a Wireless Sensor Network (WSN), weather information from a nearby Weather Station and recorded energy use of appliances and lighting fixtures. The energy appliances data was obtained by continuously measuring (every 10 minutes) a low-energy house in Belgium for 137 days (around 4.5 months). The data set contains 19,735 instances, including 26 explanatory variables and 1 temporal variable (date/time). Figure~\ref{model} shows all the variables.

In order to check which time series in the data set are non-stationary, we use the Augmented Dickey-Fuller (ADF) \cite{adf} and Kwiatkowski-Phillips-Schmidt-Shin (KPSS) \cite{kpss} tests with a confidence level of 95\%. ADF is used to determine the presence of unit root in the series and KPSS is used to check for stationarity of a time series around a deterministic trend. 

Both tests conclude that the \emph{Temperature in ironing room} and \emph{Temperature in parents room} series are non-stationary, while \emph{Appliances energy consumption} series is stationary. According to KPSS, the other series are difference stationary -- one differencing is required to make the series stationary. Therefore, in the data set, there are non-stationary times series.

Since the appliances energy consumption (Wh) measured is the focus of our analysis, it was chosen as the target variable $V^*$ and the set of explanatory variables $V$ is composed by 26 variables. 

In this work, the number of fuzzy sets $\kappa$ is 5, the length of the residuals window $w_e$ is 3 and the kernel coefficient of KPCA $\gamma$ is 0.1. We select these parameters using a grid search where we tested the following parameters (Table~\ref{parameters}) in PCA-NSFTS and KPCA-NSFTS models:

\begin{table}[h]
\centering
\caption{Parameters values tested in the grid search}
\begin{tabular}{|l|l|ll} 
\cline{1-2}
\multicolumn{1}{|c|}{\textbf{Parameter}} & \multicolumn{1}{c|}{\textbf{Values}} &  &   \\ 
\cline{1-2}
Number of fuzzy sets                 & 5, 15, 30, 45, 60                        &  &   \\ 
\cline{1-2}
Length of the residuals window                                 & 3, 4, 5                            &  &   \\ 
\cline{1-2}
Kernel coefficient                                & 0.1, 10, 0.5                            &  &   \\ 
\cline{1-2}
\multicolumn{1}{l}{}                     & \multicolumn{1}{l}{}                 &  &  
\end{tabular}
\label{parameters}
\end{table}

We divided 75\% of data for training set and 25\% for testing and compute the accuracy metrics (Subsection~\ref{experiments_methodology}) over the test set for each parameter combination. The result showed that the highest accuracy is achieved using the parameter values presented above and the accuracy are controlled by $\kappa$ and $w_e$.

Since the parameter adaptation procedure updates the position and length of the fuzzy sets constantly, we could avoid that the model generate underfitting due the small number of fuzzy sets and predict the appliances energy consumption accurately.


\subsection{Experiments methodology}
\label{experiments_methodology}

In this work, we separate 75\% of data for training set and 25\% for testing and we use the sliding window cross-validation in the computational experiments. The sliding window is a re-sampling procedure based on splitting the data set into more than one training and test subsets. The overall prediction accuracy is obtained by looking at the metrics measures over all the testing subsets. 


The 19,735 instances of the data set were divided in 30 data windows with 657 instances. For each window, we train the proposed models (PCA-NSFTS and KPCA-NSFTS) using the training set, apply the model to the test set and compute forecasting metrics over the test set. Thus, each model has 30 experiments and we evaluate the performance of ENSFTS from the average error value measured in all windows used for forecasting in the experiments. 

The following standard accuracy metrics were used: the root mean squared error (RMSE), the coefficient of determination ($R^2$), the mean absolute error (MAE) and the mean absolute percentage error (MAPE). These metrics were used to evaluate the performance of the proposed approach against the competitor models: MLR, SVM radial, GBM, RF, \cite{dataset_appliance}, MLP \cite{energy_prediction_mlp}, KNN, ERF, LSTM \cite{energy_prediction_lstm} and persistence/naive, which is a reference technique that assumes that $y(t)$ equals $y(t-1)$. 

In addition to the performance evaluation indices presented above, we evaluate the performance of the ENSFTS using the skill score index. The skill score defines the difference between the forecast and the reference forecast. The skill score can be also applied not only for comparison with a naive model but also for inter-comparisons of different forecasting methods \cite{MachineLearningSolarRadiationForecasting}. For example, a skill score equal to 0.50 means an improvement in a accuracy metric of 50\% with respect to the competitor model. A negative value indicates a performance that is worse than the competitor.

\begin{equation} \label{eq:skillScore}
Skill Score = 1 - \frac{Metric_{forecasted}}{Metric_{reference}}
\end{equation}

The ENSFTS was implemented and tested using the programming language Python 3 and the open-source pyFTS \cite{pyFTS}  and scikit-learn \cite{scikit-learn} libraries.

\section{Results}
\label{results}

Table~\ref{metrics} presents the results of RMSE, MAE, MAPE and $R^2$ for each competitor model with all the features and feature selection, as well as the accuracy metrics results for PCA-NSFTS and KPCA-NSFTS proposed models. Comparing the results with those obtained by competitors, it is clear that PCA-NSFTS and KPCA-NSFTS outperform them. Besides, KPCA-NSFTS is just slightly superior than PCA-NSFTS in all the accuracy metrics, but not significantly. This is evidence of presence of  linear correlations among variables in the data set.   

\begin{table}[h]
\centering
\caption{Model performance in the testing set (FS = feature selection)}
\begin{tabular}{|l|c|c|c|c|} 
\hline
\rowcolor[rgb]{0.753,0.753,0.753} \multicolumn{1}{|c|}{\textbf{Model}} & \textbf{RMSE (\%)} & \textbf{MAE (\%)} & \textbf{MAPE (\%)} & \textbf{$\mathbf{R^2}$ (\%)}  \\ 
\hline
MLR   \cite{dataset_appliance}                                                             & 93.18              & 51.97             & 59.93              & 16               \\ 
\hline
SVM radial  \cite{dataset_appliance}                                                             & 70.74              & 31.36             & 29.76              & 52               \\ 
\hline
GBM \cite{dataset_appliance}                                                                     & 66.65              & 35.22             & 38.29              & 57               \\ 
\hline
GBM (FS) \cite{dataset_appliance}                                                                     & 66.21              & 35.24             & 38.65              & 58               \\ 
\hline
RF \cite{dataset_appliance}                                                                      & 68.48              & 31.85             & 31.39              & 57               \\ 
\hline
MLP \cite{energy_prediction_mlp}                                                                  & 66.29              & 29.55             & 27.29              & 56               \\ 
\hline
MLP (FS) \cite{energy_prediction_mlp}                                                                  & 59.84              & 27.28             & 27.09              & 64               \\ 
\hline
KNN (FS)  \cite{energy_prediction_lstm}                                                                 & 64.99              & -                 & -                  & 58               \\ 
\hline
ERF (FS)  \cite{energy_prediction_lstm}                                                                   & 59.81              & -                 & -                  & 64               \\ 
\hline
LSTM (FS) \cite{energy_prediction_lstm}                                                                   & 21.36              & -                 & -                  & 97               \\ 
\hline
Persistence                                                            & 64.74              & 29.10             & 24.82              & 40               \\ 
\hline
\textbf{PCA-NSFTS}                                                     & \textbf{11.89}     & \textbf{3.17}     & \textbf{3.67}      & \textbf{98}      \\ 
\hline
\textbf{KPCA-NSFTS}                                                    & \textbf{11.52}     & \textbf{2.68}     & \textbf{2.91}      & \textbf{98}      \\
\hline
\end{tabular}
\label{metrics}
\end{table}

Table~\ref{skillscore} shows the skill score of PCA-NSFTS and KPCA-NSFTS with respect to some competitor models. The accuracy metric selected was the RMSE. 

\begin{table}
\centering
\caption{Skill score of PCA-NSFTS and KPCA-NSFTS}
\label{skillscore}
\begin{tabular}{|l|c|c|} 
\hline
\rowcolor[rgb]{0.753,0.753,0.753} \diagbox{\textbf{Competitor}}{\textbf{ENSFTS}} & \multicolumn{1}{l|}{\textbf{PCA-NSFTS}} & \multicolumn{1}{l|}{\textbf{KPCA-NSFTS}}  \\ 
\hline
Persistence                                                                      & 0.81                                    & 0.82                                      \\ 
\hline
GBM (FS)                                                                         & 0.82                                    & 0.82                                      \\ 
\hline
MLP (FS)                                                                         & 0.80                                    & 0.80                                      \\ 
\hline
LSTM (FS)                                                                        & 0.44                                    & 0.46                                      \\
\hline
\end{tabular}
\end{table}

PCA-NSFTS presented an improvement in RMSE by 81\% with respect to persistence model. In relation to GMB, the improvement is 82\% and PCA-NSFTS showed an enhancement of 80\% compared to MLP. KPCA-NSFTS had an improvement in RMSE by 82\% with respect to Persistence and GBM. In regard to MLP, the  enhancement is 80\%.  The best model among the competitors was LSTM, which is a state-of-the-art deep-learning method. Compared to LSTM, PCA-NSFTS has an improvement in RMSE by 44\% and KPCA-NSFTS presented an enhancement in RMSE by 46\%. 

It can be seen from the results above that, compared to competitors models, PCA-NSFTS and KPCA-NSFTS achieve optimal prediction performance on appliances energy consumption data set.

The embedding techniques allow us to extract and exploit a new feature space that better represents the inherent complexity of multivariate time series, also mitigating collinearity phenomena and catching latent interactions among features. Both PCA and KPCA algorithms can be used to identify the main component in the appliance energy consumption based on available historical data. The FTS learning approach allows us to handle non-stationary time series as well as scenarios with concept drift accurately.

\section{Conclusions}
\label{conclusions}

In this work, we investigated the possible benefits provided by a method that combines embedding transformation and fuzzy time series forecasting approach for tackling the concept drift events in multivariate time series data. We proposed a new approach for tackling high-dimensional non-stationary data, applying data embedding transformation and FTS models.

The proposed approach (ENSFTS) aimed to address the appliances energy consumption forecasting problem. The PCA and KPCA algorithms were used to extract new feature space that better represents the content of appliance energy consumption multivariate time series for the subsequent forecasting task. The embedding methods allow us to extract the relevant information that supports the target variable forecasting. 

Our experimental evaluation showed that, compared to other state-of-the-art forecasting methods, ENSFTS achieves the best prediction performance on appliances energy consumption problem. Therefore, our approach has a great value in smart home IoT applications, and can help homeowners reduce their power consumption and provides better energy-saving strategies. Besides, the proposed approach generates forecasting models readable and explainable and their accuracy are controlled basically by two parameters: the partitioning of the target variable (number of fuzzy sets) and the length of the residuals window.


\bibliographystyle{IEEEtran}  
\bibliography{references} 

\end{document}